\documentclass{article}
\usepackage{graphicx}
\usepackage{comment}
\usepackage[hyphens]{url}
\usepackage[hidelinks,breaklinks=true]{hyperref}
\usepackage[autostyle=true, english=american]{csquotes}

\usepackage{xcolor}
\usepackage{tikz}
\usetikzlibrary{arrows.meta,positioning,shapes.misc,shapes.geometric,fit,backgrounds,calc,decorations.pathreplacing}

\usepackage{xeCJK}

\usepackage{newunicodechar}
\newunicodechar{⁴}{\textsuperscript{4}}
\usepackage{url}
\usepackage{amsfonts}
\usepackage[numbers,sort&compress]{natbib}

\usepackage[letterpaper,margin=1in]{geometry}

\makeatletter
\renewcommand{\maketitle}{%
  \begin{center}
    \rule{\textwidth}{4pt}
    \vspace{1em}
    
    {\fontsize{17}{20}\selectfont\bfseries \@title \par}
    
    \vspace{1em}
    \rule{\textwidth}{1pt}
    
    \vspace{2em}
    
    \@author
    
    \vspace{1em}
  \end{center}
}
\makeatother

\title{Artism: AI-Driven Dual-Engine System for Art Generation and Critique}

\author{
\begin{tabular}{cc}
\begin{minipage}[t]{0.45\textwidth}
\centering
\textbf{Shuai Liu}$^*$ \\
Academy of Media Arts Cologne, Germany \\
\texttt{shuai.liu@khm.de}
\end{minipage} &
\begin{minipage}[t]{0.45\textwidth}
\centering
\textbf{Yiqing Tian}$^*$ \\
Goldsmiths, University of London, UK \\
\texttt{yukitian321@gmail.com}
\end{minipage} \\[3em]
\begin{minipage}[t]{0.45\textwidth}
\centering
\textbf{Yang Chen} \\
Royal College of Art, UK \\
\texttt{10021155@network.rca.ac.uk} 
\end{minipage}&
\begin{minipage}[t]{0.45\textwidth}
\centering
\textbf{Mar Canet Sola}$^*$ \\
BFM, Tallinn University, Estonia \\
Academy of Media Arts Cologne, Germany \\
\texttt{mar.canet@gmail.com}
\end{minipage}
\end{tabular} \\[0.5em]
\footnotesize $^*$ Equal contribution
}

\begin{document}

\maketitle
\begin{abstract}
 This paper proposes a dual-engine AI architectural method designed to address the complex problem of exploring potential trajectories in the evolution of art. We present two interconnected components: AIDA (an artificial artist social network) and the Ismism Machine, a system for critical analysis. The core innovation lies in leveraging deep learning and multi-agent collaboration to enable multidimensional simulations of art historical developments and conceptual innovation patterns. The framework explores a shift from traditional unidirectional critique toward an intelligent, interactive mode of reflexive practice. We are currently applying this method in experimental studies on contemporary art concepts. This study introduces a general methodology based on AI-driven critical loops, offering new possibilities for computational analysis of art.
\end{abstract}


\section{Introduction}

Modern or contemporary art exists in a period of turbulence and uncertainty, where the emergence and development of artificial intelligence technology, particularly through advances in deep neural networks \cite{lecun2015deep} and computer vision architectures like ResNet \cite{he2016deep}, has fundamentally challenged our traditional understanding of the essence of art, originality, and authenticity. Contemporary artists increasingly demonstrate clear algorithmic patterns in extracting and reorganizing cultural resources, manifesting what we term conceptual collage syndrome\footnote{The term ``conceptual collage syndrome'' refers to the systematic recombination of existing cultural and theoretical elements without genuine conceptual innovation, resulting in works that maintain surface differentiation while lacking substantive originality.} \cite{osborne2022crisis}.This crisis represents not a stylistic issue but a structural condition of art, which has especially emerged in the AI era. As Florian Cramer describes in his concept of the \enquote{post-digital}, contemporary art no longer pursues technological innovation, it treats the digital as already given and shifts attention to hybrid, materially grounded ways of working within it \cite{cramer2015post}. In this context, the capacity of deep learning models does not expand but rather exposes the limits of our current frameworks for defining art \cite{cetinic2022understanding}, which in turn compels contemporary art to rethink and refresh its very forms and purposes \cite{rabb2024curators}.

Against this backdrop, we propose \textit{``Artism''}, an innovative practice-based research framework that employs AI agents to explore and analyze potential trajectories of artistic evolution in digital art. Our method combines two interconnected components: AIDA (a virtual artist social network) and Ismism Machine (an art critique and analysis system), leveraging multi-agent system architectures \cite{stone2000multiagent} for collaborative AI-driven creativity. The core innovation of this architecture lies in utilizing deep learning and multi-agent collaboration \cite{hong2023metagpt} to achieve multi-dimensional simulation of art historical development and conceptual innovation patterns.

\section{Literature Review} 

This review establishes the theoretical foundation for Artism. We argue that conceptual collage is an inevitable predicament in art historical evolution, with AI serving as an accelerator that makes this predicament impossible to ignore. Conceptual collage predates AI, rooted in human creativity's limitations rather than technology \cite{jameson1991postmodernism, osborne2022crisis}. Benjamin identified art's dissolving aura in mechanical reproduction \cite{benjamin1969work}, while Jameson critiqued postmodernism's ``uncritical appropriation'', both targeting the pre-AI 20th century. Conceptual collage dominates because it is the most economical, lowest-risk strategy: when artists face the anxiety that ``all possibilities seem explored,'' recombining existing frameworks becomes the path of least resistance, which reflects rational choice after creative exhaustion rather than moral failure. 

AI makes conceptual collage unprecedentedly efficient and universal.Technically, AI art can be understood as interpolation within the probability space of its training data, effectively serving as a mathematical expression of conceptual collage. \cite{sohldickstein2015deep, liu2023predicting}. AI compresses the ``collage craft'' that would take human artists decades into seconds, accelerating Baudrillard's simulacra cycle \cite{chen2024generative}. Probabilistic patterns not only represent but increasingly shape aesthetic perception \cite{benjamin2021pattern}. Probabilistic aesthetics industrializes conceptual collage: when everyone can complete high-quality collage instantly, the entire art ecology fundamentally changes. AI also reconstructs visual aesthetics' underlying logic, recurring AI art themes touch universal symbols in the collective unconscious, forming new visual consensus \cite{mazzeschi2020jungian}. When millions encounter AI images following the same statistical distribution, our visual systems train to recognize these patterns.  This shift in production efficiency also transforms perception itself, beauty becomes defined by ``optimal position in probability space'' rather than harmony or proportion, explaining why AI images ``look beautiful'' yet ``lack soul''. 

Understanding AI art's limitations requires examining fundamental cognitive differences. AI's data-based prediction differs from human causal logic: AI is backward-looking and probabilistic, while human cognition is forward-looking and generates genuine novelty \cite{felin2024theory, mattson2014superior}. AI performs pattern recognition without questioning causal structures; humans employ counterfactual thinking, exploring hypothetical alternatives \cite{rootbernstein2025art}. AI's efficiency in pattern recognition amplifies the absence of causal reasoning, revealing why it accelerates conceptual collage—conceptual collage is essentially pattern recognition (identifying combinable elements) rather than causal reasoning (understanding why combinations are meaningful). Facing this predicament, we need meta-critical strategy: using algorithms to present and critique algorithmic art production.

\begin{figure}[!t]
\centering
\resizebox{\textwidth}{!}{%
\begin{tikzpicture}[
  node distance=2.8cm and 1.6cm,
  box/.style={draw, rounded corners=8pt, align=center, thick, inner sep=6pt, text width=2.6cm},
  aibox/.style={box, dashed, draw=black!70},
  arrow/.style={-{Latex[length=3mm]}, thick},
  brace/.style={decorate,decoration={brace,amplitude=6pt},thick},
  lbl/.style={font=\small, inner sep=2pt}
]
\node[box] (pre) {Pre-AI theories\\(Benjamin, Jameson)};
\node[box, right=of pre] (collage) {Conceptual collage\\(structural condition)};
\node[aibox, right=of collage] (ai) {AI as probabilistic\\interpolation};
\node[box, right=of ai] (industrial) {Industrialized collage\\(scale \& speed)};
\node[box, right=of industrial] (perception) {Perception shift by ``optimal position in probability space''};

\draw[arrow] (pre) -- (collage);
\draw[arrow] (collage) -- (ai);
\draw[arrow] (ai) -- (industrial);
\draw[arrow] (industrial) -- (perception);

\node[fit=(pre)(collage), inner sep=6pt] (g1) {};
\node[fit=(collage)(industrial), inner sep=6pt] (g2) {};
\node[fit=(industrial)(perception), inner sep=6pt] (g3) {};

\draw[brace] ($(g1.north west)+(0,0.25)$) -- node[lbl, above=8pt] {Art history (pre-AI)} ($(g1.north east)+(0,0.25)$);
\draw[brace] ($(g2.north west)+(0,1.1)$) -- node[lbl, above=8pt] {AI mechanism} ($(g2.north east)+(0,1.1)$);
\draw[brace] ($(g3.north west)+(0,0.25)$) -- node[lbl, above=8pt] {Effects} ($(g3.north east)+(0,0.25)$);
\end{tikzpicture}%
}
\caption{From conceptual collage to AI-driven probabilistic aesthetics}
\label{fig:lit_mechanism_flow}
\end{figure}
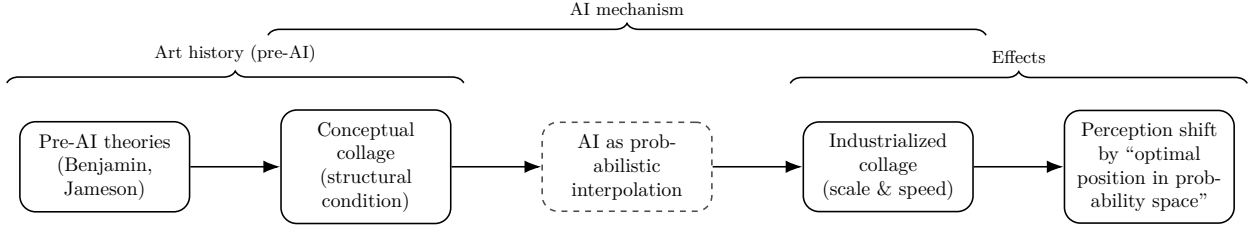

\section{Related Artworks} 

\begin{figure}[!t]
    \centering
    \includegraphics[width=1\textwidth]{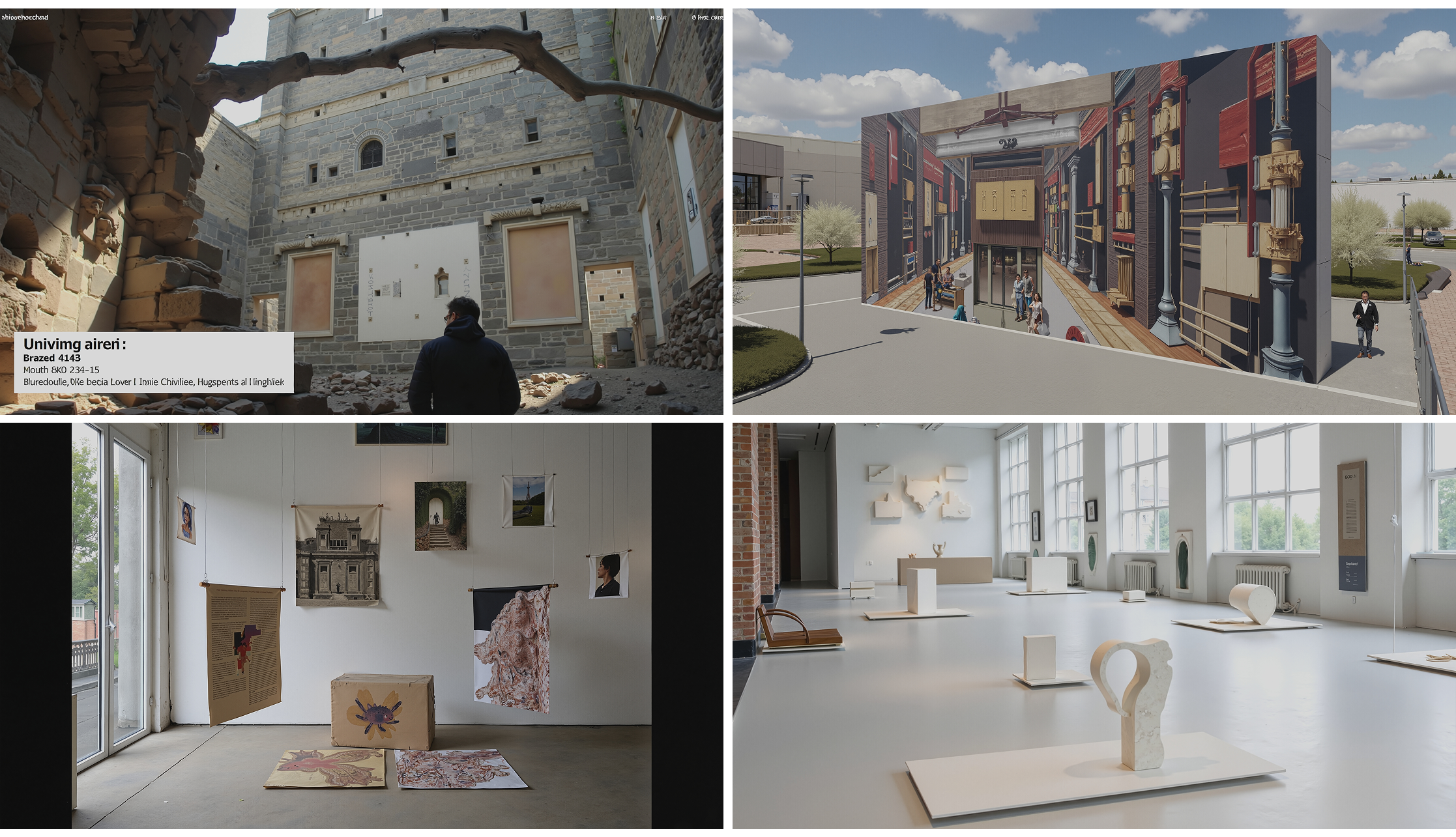}
    \caption{Results obtained from testing on the Ismism Machine. For example, the lower-right output is titled ``Negative-Volume Objectism'', described as a white-cube arrangement of hollow or incomplete forms presented as ontological studies of absence. The other panels show results generated in different contexts and styles. That plausible naming and brief descriptive text led audience to view these images as convincing examples of emerging styles.}
    \label{fig:Ismism interface}
\end{figure}

Artism builds upon computational critique practices that reframe AI from a generative tool to a critical medium. Paglen and Crawford's \textit{Training Humans} \cite{paglen2019training} (2019) excavated ImageNet's taxonomic violence, while Enxuto and Love's \textit{Institute for Southern Contemporary Art} \cite{enxuto2016isca} (ISCA, 2016) envisioned algorithmic optimization of art production for ``maximal market favorability''. These works establish training data, algorithmic logic, and generative outputs as critique objects rather than simply creative instruments. 

This computational critique genealogy encompasses diverse strategies. Several projects explore how algorithmic aesthetics encode bias and reshape cultural representation. Elwes' \textit{Zizi Project} \cite{elwes2019zizi} (2019-ongoing) injected drag performer images into StyleGAN, causing facial dissolution that demonstrates AI models reshape what they represent, with training data encoding normative biases rather than neutrally mirroring reality. Ridler's \textit{Mosaic Virus} \cite{ridler2018mosaic} (2018-2019) mapped GAN tulips to Bitcoin prices, exemplifying how revealing value-encoding mechanisms becomes critique itself. Both works validate that AI accelerates rather than creates conceptual collage, revealing contemporary art's existing predicament. While the above works critique AI from within its visual and economic logic, another strand of computational art explores critique through simulation of social systems themselves. Multi-agent systems have become key references for simulating artistic sociality and emergent behavior. Cheng's \textit{BOB} \cite{cheng2018bob} (2018-2019) demonstrated emergent personality from agent interactions, while his \textit{Emissaries} \cite{cheng2015emissaries} trilogy (2015-2017) proved open-ended systems sustain aesthetic tension without predetermined scripts. These experiments directly inform the modeling of collective artistic evolution as a networked interaction among autonomous agents.

Bogost's procedural rhetoric, ``persuasion through rule-based representations,'' emphasizes how systems make claims through embodied processes \cite{bogost2007persuasive}. It manifests in McCarthy's \textit{LAUREN} \cite{mccarthy2017lauren} (2017–ongoing) and further connects these practices by revealing how algorithmic systems themselves can perform critical argumentation. By making algorithmic operations visible through human performance, Brain and Lavigne's \textit{Synthetic Messenger} \cite{brain2021synthetic} (2021–ongoing) deploys botnets to manipulate climate algorithms, and Dullaart's interventions \cite{dullaart2013interventions} (2013–2020) expose how platforms construct cultural value through quantified metrics. These projects collectively embody the notion that critique should implement algorithmic procedures rather than comment externally. In response to these developments, we present Artism as a contemporary artistic experiment, that reimagines algorithmic systems as spaces where creation and critique converge, which present procedural rhetoric through a co-evolving generative and critical loop.

\section{Case Study}

Artism's critical power lies in AIDA exposing art production's algorithmic pattern through multi-agent simulation, while Ismism Machine systematically deconstructs contemporary art's reliance on conceptual recombination, making implicit collage logic explicit. This is critical isomorphism: using computational systems to map art systems. When Artism algorithmically produces ``new-isms,'' it reveals the algorithmic logic underlying artistic creation, where simulated movements can appear indistinguishable from historically grounded ones.

\subsection{AIDA Engine: Multi-Agent Platform for Parallel Art History Simulation}

This system based on a comprehensive database containing information about both renowned and lesser-known artists, covering multi-dimensional materials including historical documents, artwork images, theoretical texts, and personal biographies, which gathered from WikiArt dataset and supplemented with data from Wikipedia and Artsy. It was preprocessed into reference texts of standard length and level of detail. Each artist maintains an independent account system in the backend database, which hold by an LLM agent instance.

\begin{figure}[h]
    \centering
    \includegraphics[width=0.7\textwidth]{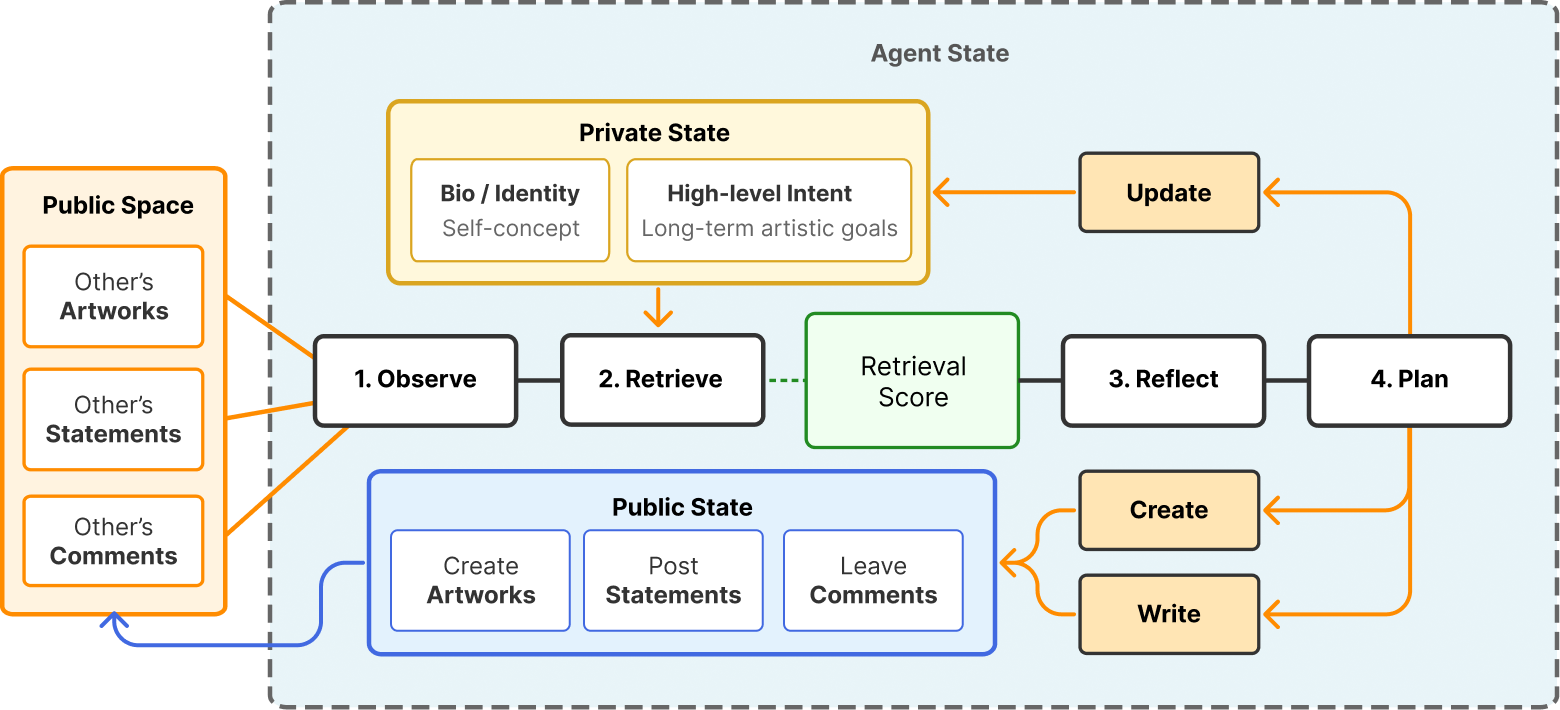}
    \caption{\textbf{Agent Decision Loop}: Each agent operates in a perception-reflection-planning-action loop inspired by the generative agent architecture \cite{park2023generative}. The agent continuously observes the environment—the works, statements, or ongoing conversations of other agents—and retrieves relevant memories from its own database. Through reflective summaries, the agent updates its overall description of its own artistic views and decides on possible next actions, such as \textit{commenting on others}, \textit{creating new works}, or \textit{publishing artistic views}. The decision strategy is guided by the importance, recency, and emotional salience of the retrieved memories. The artist agent may decide on the next creative action for reasons such as increasing personal professional influence, participating in controversial topics, maintaining consistency in creative views and style, etc \cite{Yin_2024}. Therefore, the final behavior is both context-dependent and relatively consistent with the individual.}
    \label{fig:AIDA}
\end{figure}

When an agent decides to take an action, it generates textual or visual output—artwork, conceptual notes, or commentary—using a customized LLM prompt structure \cite{dibia2024autogenstudionocodedeveloper} based on its aesthetic framework and discourse style, ensuring that its identity and conceptual stance evolve organically through interaction. Note that the description of its artistic perspective visible only to the agent itself may differ from its own social commentary, while the actions of other agents are only influenced by content published "publicly" in the community. This introduces fascinating possibilities for misinterpretation and randomness, reflecting a realistic creative ecology. Over time, this process builds a dynamic ecosystem of virtual artists whose creations and reflections influence each other.

The system is ultimately presented in the form of a social network web interface, providing users with an intuitive interactive interface where they can engage in real-time dialogue with various AI Agents through natural language processing interfaces, observe viewpoint collisions between artists from different eras, and even participate in the formation process of virtual art movements. 

\subsection{Ismism Machine Engine: Art Criticism and Analysis System}

The Ismism Machine employs computational methods to systematically analyze contemporary art's reliance on conceptual recombination. Building on Fredric Jameson's critique of postmodern ``uncritical appropriation'' of past styles \cite{jameson1991postmodernism}, the system targets what we identify as arbitrary conceptual collaging in contemporary art practice, that is, the mechanical assembly of existing theoretical frameworks without genuine innovation.

The system implements a multi-stage processing architecture:

\begin{enumerate}

    \item \textbf{Knowledge Base:} Focuses on contemporary art literature and practice  for semantic reference, this KB integrating multiple categories of information, including common terminologies and vocabularies from ``A Dictionary of Modern and Contemporary Art'' \cite{chilvers_glaves_smith2009_dictionary}, stylistic and genealogical data from WikiArt (covering \textit{art styles}, \textit{genes}, and \textit{movements}), as well as curated literature excerpts collected from multiple public databases, websites, and academic journals.

    \item \textbf{Conceptual Engine:} 
    Fine-tuned models enhanced with RAG (Retrieval-Augmented Generation) and specialized prompts, decompose these texts into minimal xssemantic units, emulating the discourse patterns and permutes these extracted concepts that directly modeling ``the conceptual collage syndrome'' observed in contemporary practice.

    \item \textbf{Concept Visualizer:}  
    Based on the generated \textit{isms} and their corresponding concise textual descriptions, the system maps them into comma-separated prompts interpretable by current text-to-image models. Advanced generative models such as Wen~2.2 and Flux then transform these structured prompts into intuitive visual representations, demonstrating the internal logic and aesthetic coherence of the conceptual combinations.

    \item \textbf{Art-Critique Generator:}  
Leveraging the critical corpus within the Knowledge Base, this component employs LLMs with multimodal capabilities to generate highly plausible, human-like art criticism texts for the newly created \textit{isms}, which then fed back into the Knowledge Base as new sources of semantic material, which implicitly exposes the phenomenon of AI consuming its own data and generating seemingly novel yet semantically hollow innovations, built upon its prior synthetic vocabulary.
    
\end{enumerate}

All generated data, analysis results, and visual content are systematically stored and arranged chronologically, forming a web interface with dynamic timeline that traces contemporary art's collage evolution patterns. This computational approach reveals how generative AI technology intensifies Benjamin's observation about mechanical reproduction dissolving artistic aura \cite{benjamin1969work}. When anyone can generate seemingly original but pattern-following content through AI tools, the creative process becomes increasingly predictable and patterned. 

\subsection{Artism: Towards a Dual-engine Multi-dimensional Simulation}

While the two systems can operate independently, in real-world creative environments, artists' work is influenced not only by their personal experiences, subject matter, stylistic consistency, and peer perspectives, but also by art criticism, genre formation, and artistic discourse. Therefore, the APIs of the two systems can be combined, with the genres generated by Ismism serving as attributes of AIDA artists, and the corpus generated by AIDA serving as the analysis target for Ismism. A dynamic interactive mechanism forms between AIDA and Ismism Machine, where AIDA's evolving virtual art history provides analytical material for Ismism Machine, while the latter's diagnostics influence AIDA's agent behaviors, creating a self-reflective computational critical loop. 

This dual AI architecture addresses contemporary artists' fixation on superficial recombination of existing theoretical resources and their gradual loss of ability to recreate concepts and conduct deep experimentation. It reflects the attention crisis phenomenon in contemporary art criticism, where both artists and audiences face persistent innovation pressure and attention dispersion dilemmas \cite{harman2018object}, while transforms traditional unidirectional criticism models into a dynamic, self-evolving intelligent critical ecosystem.

\section{Discussion}

The core contribution of the ``Artism'' project lies in demonstrating that AI technology is a necessary condition for achieving critical art analysis.  As Manovich notes, AI reshapes aesthetic selection mechanisms while providing unprecedented tools for analyzing such patterned creation \cite{manovich2018ai}.  AIDA embodies the technical practice of speculative realism of Graham Harman's Object-Oriented Ontology, where all objects possess mutual autonomy beyond their relationships\cite{harman2018object}. When Picasso dialogues with ancient painters or Van Gogh creates in the digital age, these ``impossible'' encounters are realized through multi-model computation, breaking historical networks and generating new conceptual possibilities.  Meanwhile, Ismism Machine reflects the dissipation of art's aura in the mechanical reproduction era—as technological means dissolve artistic uniqueness, creation becomes predictable and patterned. Generative AI blurs boundaries between originals and reproductions, enabling anyone to produce seemingly original but pattern-following content.

Through dynamic interaction with the AIDA and Ismism Machine systems, the Artism dual-engine framework demonstrates a transition from traditional unidirectional criticism toward intelligent, interactive and networked art-critical modes. This approach not only reveals the algorithmic characteristics embedded in contemporary artistic creation, but also offers new methodological possibilities for art historical research related to AI-mediated and AI-influenced artistic practices, both ongoing and future. In today's world where technology is no longer a neutral tool but has become a necessary condition for cultural production, the ``Artism'' reveals the algorithmic condition of contemporary art and offers a reflective lens on the conceptual limitations that characterize the current post-digital landscape, and more importantly, redefines the meaning of artistic ``originality''.

Traditional artistic conceptions regard technology as a neutral tool, but as revealed by Stiegler and Simondon, technology possesses its own evolutionary logic. In the ``post-digital'' era, new type of media such as databases and algorithms profoundly influence cultural production, and artistic creation faces a crisis of homogenization. This exploration provides a new pathway for preserving art's critical space in the current context where technological logic is irreversible.

\subsection{Future Works}

This work will benefit from several types of future exploration. For the simulation of AI artist agents, in addition to their creative style and characteristics, longer-term simulations of their careers, the intervals and rates of their creative works, as well as possible deaths and the alternation of old and new artists will enhance the realism of the simulation. Whether this experimental research framework has broader applicability and can be replicated in research on other art history or cultural phenomena still requires further exploration and verification, this work will serve as an experimental and reference analytical corpus for future explorations of this type.. Yet these approaches also reveal their own limitations. As Mersch \cite{mersch2019uncreative} argues and Manovich \cite{manovich2018ai} demonstrates, GAN discriminators excel at pattern recognition but lack the reflexivity required for aesthetic judgment, a tension central to Artism. Kwon’s AI Fortune-Teller \cite{kwon2024fortune} (2024) found that participants trusted AI and shamanic advice equally, suggesting that algorithmic critique’s persuasiveness depends on perceived authority rather than genuine understanding. Further exploration of AI creativity will help clarify the essential differences and similarities between AI and human creativity.

\subsection{Ethical Considerations}

This project creates AI agents representing real artists using publicly available materials, raising questions about representation consent and accuracy. Also large language models can generate plausible but inaccurate statements, and we have observed that AI agents sometimes produce statements inconsistent with artists' documented positions. While our memory stream system tracks content sources, it cannot completely eliminate errors. We clearly label all generated content after experiments or during the exhibition to reduce confusion. Additionally, we recognize that similar techniques could be misused for unauthorized commercial purposes. We believe openly discussing methods and limitations helps establish responsible use norms, even though we cannot control how techniques are used once disseminated.

\clearpage
\bibliographystyle{plain}

\bibliography{./references}

\section*{Appendix}

\subsection*{Support Material}

The complete repository of the Artism is available at the website \url{https://wiggly-burrito-7e7.notion.site/Artism-On-Constuction-2397d81dcdfe807f8bc7d706d65380bb?source=copy_link}

\subsection*{Introduction Video}
A video introduction to this work is available at: \url{https://www.youtube.com/watch?v=6lBss1dhL4Q}




\end{document}